\documentclass[10pt]{article} %
\usepackage[preprint]{rlc}
\usepackage{wrapfig}

\usepackage{amssymb}            %
\usepackage{mathtools}          %
\usepackage{mathrsfs}           %
\usepackage{graphicx}           %
\usepackage{subcaption}         %
\usepackage[space]{grffile}     %
\usepackage{url}                %
\usepackage{subcaption}

\usepackage{amsmath} %
\usepackage{amsthm}  %
\usepackage{thmtools}
\usepackage{thm-restate}
\usepackage{xcolor}
\usepackage{xspace}

\theoremstyle{definition} %
\newtheorem{definition}{Definition}[section]
\usepackage{thmtools, thm-restate}
 
\newtheorem{theorem}{Theorem}

\usepackage[textsize=tiny]{todonotes}

\newcommand{\Xc}{\mathcal{X}}
\newcommand{\Hc}{\mathcal{H}}
\newcommand{\Lc}{\mathcal{L}}
\newcommand{\Uc}{\mathcal{U}}

\def\Pr{\mathbb{P}}
\def\E{\mathbb{E}}
\def\F{\mathbb{F}}
\def\H{\mathbb{H}}
\def\I{\mathbb{I}}

\def\KL{\mathbf{d}_{\mathrm{KL}}}

\title{The Need for a Big World Simulator: \\
A Scientific Challenge for Continual Learning}

\author{Saurabh Kumar\footnote{Equal contribution.}  \\
    szk@stanford.edu \\
    Department of Computer Science\\
    Stanford University \\
    $^*$Denotes equal contribution.
    \hspace{-3cm}
    \And
    Hong Jun Jeon$^*$ \\
    hjjeon@stanford.edu\\
    Department of Computer Science\\
    Stanford University
    \And
    Alex Lewandowski \\
    lewandowski@ualberta.ca \\
    Department of Computing Science \\
    University of Alberta
    \hspace{5.9cm}
    \And
    Benjamin Van Roy \\
    bvr@stanford.edu \\
    Department of Electrical Engineering \\
    Stanford University \\
}

\begin{document}

\maketitle

\begin{abstract}
The ``small agent, big world'' frame offers a conceptual view that motivates the need for continual learning.  The idea is that a small agent operating in a much bigger world cannot store all information that the world has to offer. To perform well, the agent must be carefully designed to ingest, retain, and eject the right information. 
To enable the development of performant continual learning agents, a number of synthetic environments have been proposed. However, these benchmarks suffer from limitations, including unnatural distribution shifts and a lack of fidelity to the ``small agent, big world'' framing.
This paper aims to formalize two desiderata for the design of future simulated environments. These two criteria aim to reflect the objectives and complexity of continual learning in practical settings while enabling rapid prototyping of algorithms on a smaller scale.
\end{abstract}

\section{Introduction}

The real world is an unfathomably complex system, both to humanity and to the agents that it designs. The small agent, big world frame captures this perspective by positing that to an agent with bounded computational resources, or {\it capacity}, the world appears complex and non-stationary~\citep{dong2022simple}.  As such, in order for said bounded capacity agent to continually fruitfully engage with the world, it must continuously ingest new knowledge while selectively retaining previously acquired knowledge. We refer to this process as \emph{continual learning}.

Thus far, the study of continual learning has revolved around the discovery of phenomena which plague the naive application of algorithms designed in traditional machine learning settings.  Such phenomena include the likes of ``catastrophic forgetting'' which describes an agent's tendency to forget information from the past \citep{kirkpatrick2017overcoming, zenke17_contin}, and ``plasticity loss'' which describes an agent's inability to absorb new information with increased time \citep{dohare21_contin_backp, dohare23_maint_plast_deep_contin_learn, kumar23_maint_plast_regen_regul, lyle23_under_plast_neural_networ}. These studies have brought about the design of synthetic environments which produce artificially generated and controllable data streams. These environments are designed to stress test an agent's resilience to these phenomena.

However, if we frame the goal of continual learning to design agents which continue to engage fruitfully with the world, existing benchmarks do not capture the essence of the world. 
For instance, on current benchmarks, evaluation metrics for catastrophic forgetting measure an agent's ability to remember everything in the past, which is unlikely to be necessary to fulfill a definition of fruitful engagement with the world. Current benchmarks also suffer from a limited variation in data since they draw from datasets designed for non-continual learning settings, like MNIST, CIFAR, and ImageNet. Further, they exhibit unnatural distribution shifts, such as the abrupt change in pixel permutation seen in Permuted MNIST and the random reassignment of labels to images in Random Label CIFAR~\citep{goodfellow13_empir_inves_catas_forget_gradien}.  Such constraints do not aptly capture the gradual and often subtle evolution of data encountered in real-world settings.

Moreover, these benchmarks do not embody the small agent, big world view.  In particular, for natural models of an optimal capacity-constrained agent operating in a vastly more complex world, the agent is always able to substantially improve its performance if granted greater capacity. 
For example, performance shortfall might shrink by a factor of at least some fixed $A$ each time capacity is scaled by a factor $B$. In existing continual learning benchmarks, there are diminishing returns to increasing an agent's capacity. In particular, at sufficient scale, each successive factor $B$ increase in capacity reduces performance shortfall by an exponentially smaller rather than fixed factor.

In this paper, we emphasize the need for a ``big world simulator.''  By this we mean a new type of synthetic environment that more accurately reflects the objectives and complexity of continual learning in practical settings.  Such an environment should simultaneously retain the property of existing benchmarks that rapid prototyping of algorithms at small scale is possible.  Furthermore, we echo \cite{kumar2023continual} by suggesting a concrete objective -- that of minimizing average error over an infinite horizon subject to a capacity constraint -- which reflects an agent's ability to fruitfully engage with the world. 
By highlighting environment criteria and a clear objective, we aim to guide future research toward developing algorithms that can be evaluated at small scale while maintaining practical relevance in real-world applications.
\looseness=-1

We view this workshop as an opportunity to engage with the continual learning community in exploring the development of an environment simulator that can facilitate scientific advances in the area. This paper serves as an important part of this communication, laying out principles and concepts that can help frame the discussion.  The paper is organized as follows. In Section~\ref{sec:related_work}, we briefly review a common form of synthetic continual learning benchmark that is currently used for evaluating agents. In Section~\ref{sec:formalize}, we formalize the notion of an environment and an agent as they pertain to continual learning. In Section~\ref{sec:desiderata}, we propose a set of criteria for the design of future synthetic environments. In Section~\ref{sec:forgetting_and_plasticity}, we discuss the implications of these desiderata on evaluating forgetting and implasticity. Finally, in Section~\ref{sec:synthetic_env}, we propose an illustrative example of a big world simulator that satisfies the two criteria.
\looseness=-1

\section{A Common Recipe for Synthetic Continual Learning Benchmarks}\label{sec:related_work}

  Synthetic benchmarks are useful for rapid prototyping, enabling a researcher to evaluate an agent's capabilities with respect to an array of identified challenges \citep{Osband2020Behaviour, osband2022neural}.
  Despite three decades of active research on continual learning
  \citep{ring94_contin,thrun98_lifel}, 
  there remains comparative
  difficulty in i) identifying the challenges unique to continual learning and (ii)
  measuring progress made on previously identified challenges. 
  For example, recent work highlights that the challenge of catastrophic forgetting associated with existing continual learning benchmarks has been addressed by computationally inefficient solutions which are infeasible in most real-world scenarios \citep{prabhu2020gdumb}.
  In this section, we briefly
  discuss a commonly used recipe for creating synthetic continual learning benchmarks and its fundamental limitations.

 Most synthetic continual learning benchmarks use the following recipe: take an existing non-continual dataset and create a continual learning problem by incorporating some form of non-stationarity.
This data stream is often delineated into a
sequence of tasks, where the non-stationary is applied at specific points in
time to create a new task by transforming the observation distribution or the
target function. Within each task, the environment faced by the agent is
stationary. 
For example, in Permuted MNIST
\citep{srivastava13_compet,goodfellow13_empir_inves_catas_forget_gradien,
  kirkpatrick2017overcoming}, each task involves a fixed, distinct permutation of the
pixels of all images in the MNIST dataset, and the challenge is to classify the
digit of each image with these new inputs (See Appendix \ref{app:related}).\looseness=-1

This recipe for creating a continual learning benchmark has limitations stemming from its dependence on applying non-stationarity to a previously collected dataset.
First, the dataset is limited in its complexity to observations and classes that have been largely collected for the purposes of training an agent to convergence.
Second, the non-stationarities used are limited to artificial, unnatural and abrupt changes to the experience stream unlike the kind of structured non-stationarity of the world.
These limitations of the benchmark can result in an agent that either (i) is able to outscale the challenge of learning continually \citep{prabhu2020gdumb}, or, (ii) can no longer benefit from additional capacity.
There are certainly advantages to using the recipe: several datasets already exist, along with baselines and performance levels for comparison.
However, addressing these limitations can lead to a continual learning benchmark that provides a controllable, open-ended, and challenging experience stream from which an agent must learn continually.

\section{Formalizing the Environment and Agent}\label{sec:formalize}

Before outlining the criteria for a big world simulator, we first propose information-theoretic characterizations of an environment and an agent. 
These characterizations will help formalize the notion that the world appears complex and non-stationary to a bounded capacity agent, necessitating a trade-off between ingesting and ejecting information.

\subsection{Notation}
We begin with a review of notation and probability and information theory that we will use in this paper. We define random variables with respect to a common probability space $(\Omega, \F, \Pr)$.  For a random variable $X: \Omega \mapsto \mathcal{X}$, we use $\Pr(X\in\cdot)$ to denote the distribution of $X$.  We use $\H, \I$ to denote (conditional) entropy and mutual information respectively.  Concretely, for random variables $X,Y,Z$
$$\H(X|Y) = \E\left[ \ln \frac{1}{\Pr(X|Y)} \right];\quad \I(X;Y|Z) = \H(X|Z) - \H(X|Y,Z).$$
We use the notation $\KL$ to denote the KL divergence, which maps two probability distributions to $\Re_+ \cup \{+\infty\}$, where $\Re_{+}$ denotes the non-negative reals.

\subsection{The Environment}
Let $(\Xc, \Sigma_{\Xc})$ be a measurable space which consists of the set of observations and a sigma-algebra on that set.  
Let $\Hc = \{\Xc^n: n \in \mathbb{N}\}$ denote the set of all \emph{histories} of observations.  An environment is identified by a function $f$ which maps $\Hc$ to the set of probability measures $\Sigma_{\Xc}\mapsto[0,1]$.

We assume that the environment generates a stream of observations which is characterized by a stochastic process $(X_0, X_1, X_2, \ldots)$ defined with respect to a probability space $(\Omega, \F, \Pr)$.  The conditional distribution $\Pr(X_{t+1}\in\cdot|H_t) = f(H_t)$, where for all $t \in \mathbb{Z}_{+}$, $H_t = (X_0, X_1, \ldots, X_t) \in \mathcal{X}^n$ denotes the history of observations up to time $t$.

\subsection{The Agent}
In this work, we consider an agent to be a map $\pi: \Uc \times \Xc \mapsto \Uc$ where $\Uc$ denotes a set of \emph{agent states}.  
For all $t$, we let $U_{t} = \pi(U_{t-1}, X_t)$ denote the agent state at time $t$, which is computed from the previous agent state $U_{t-1}$ and the latest observation $X_t$. We represent constraints placed on the agent via an \emph{informational} constraint $\I(H_t; U_t) \leq c$ for all $t$.  We refer to this value $c$ as the \emph{capacity} of the agent.

\subsection{Predictions and Error}
An agent is evaluated for its ability to accurately predict future observations.  Concretely, for any agent $\pi$, we define the error of $\pi$ to be
$$\Lc_{\pi} \ =\ \limsup_{T\to\infty}\ \E_{\pi}\left[\frac{1}{T}\sum_{t=0}^{T-1}\KL\left(\underbrace{\Pr(X_{t+1}\in\cdot|H_t)}_{\rm environment\ probability}\ \|\ \underbrace{\Pr(X_{t+1}\in\cdot|U_t)}_{\rm agent\ prediction}\right)\right].$$
We characterize an agent's effectiveness via it's error.  Since the function $f$ (environment probabilities) may depend arbitrarily on $H_t$ it is clear that an agent with limited capacity will incur nonzero error.  For all $t$, $U_t$ will likely be missing information about $H_t$ which may be relevant for making predictions about $X_{t+1}$.  Therefore, an effective agent ought to selectively remember and forget aspects of $H_t$ which lead to the best predictions about the future, while obeying the capacity constraint.

\begin{wrapfigure}{r}{0.4\linewidth}
    \centering
    \vspace{-8mm}
    \includegraphics[width=0.4\textwidth]{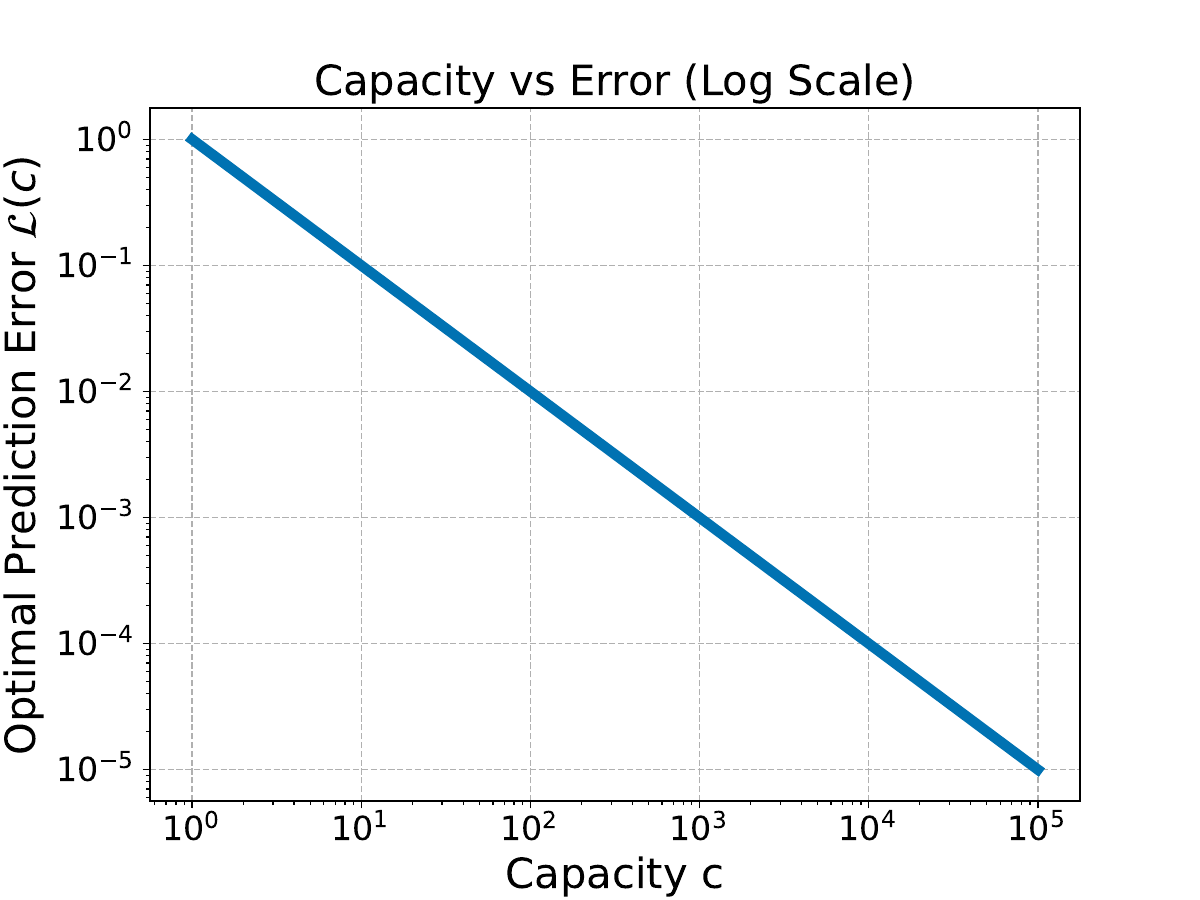}
    \caption{To illustrate the notion that there are no diminishing returns to increasing agent capacity, we plot what the capacity versus optimal prediction error would look like for a $1$-complex environment. This is simply the curve $\Lc(c) = \frac{1}{c}$. In order reduce error by a factor of $10$, we must put forward $10$ times the capacity. When both the x and y-axes are in log scale, this curve appears as a line with slope of value $1$.}
    \label{fig:capacity_vs_error}
    \vspace{-8mm}
\end{wrapfigure}

\section{Desiderata for a Big World Simulator}\label{sec:desiderata}

Equipped with formal characterizations of an environment, an agent, and error incurred by an agent interacting with the environment, we now propose two criteria for a big world simulator.  

\subsection{There are no diminishing returns to increasing an agent's capacity.}

In the small agent, big world setting, the world is enormously -- perhaps even infinitely -- complex relative to any finite-capacity agent. Since there is always more to learn about the world, a desired feature of continual learning agents is that endowing an agent with significantly increased capacity should invariably lead to substantial improvement in its capabilities. We frame this feature as a property of a big world simulator. 

To capture the notion that increasing capacity yields significant performance improvements, we establish a constraint on how the prediction error diminishes as the agent's capacity expands.  First, let
\looseness=-1
$$\Lc(c)\ =\ \inf_{\pi: \forall t,\ \I(H_t;U_t) \leq c}\ \Lc_{\pi}.$$
$\Lc(c)$ denotes the optimal error incurred by an agent with capacity $c$.  We draw inspiration from the research in large language models which observes the desired phenomenon of continued reduction in out-of-sample error for every extra unit of available compute \citep{kaplan2020scaling, hoffmann2022training}.  In these works, they establish a \emph{power law} relationship between the error and capacity.  This is in contrast to many classical statistical settings in which the reduction in out-of-sample error decays \emph{exponentially} fast in $c$.  We now present the notion of $k$-complexity:
\begin{definition}
    For $k \in \Re_{++}$, an environment is \textit{$k$-complex} if 
    $$\Lc(c) \ =\ \Theta\left(\frac{1}{c^k}\right).$$
\end{definition}
This definition characterizes an environment's ability to consistently offer performance improvements in response to increases in agent capacity. An illustration of what the capacity vs error curve would look like for a $1$-complex environment is provided in Figure~\ref{fig:capacity_vs_error}.

\subsection{An optimal finite capacity agent interacting with the environment will never stop learning.}
Since our goal is to evaluate the continual learning capabilities of agents, the environment should be designed such that an effective capacity constrained agent must \emph{never stop learning}. To formalize this concept, we begin by framing nonstationarity as a description of the agent's subjective experience.
\begin{definition}
\label{def:non-stationarity}
    An environment $f$ is \emph{non-stationary} with respect to an agent $\pi$ if
    $$\limsup_{t\to\infty}\ \I(X_{t+1};X_{t+2:\infty}|U_t) > 0.$$
\end{definition}
This states that an agent $\pi$ will always continue to encounter new information in $X_{t+1}$ which will be relevant for making predictions about the future $X_{t+2:\infty}$ beyond what is currently known in $U_t$.  Therefore, in our environment design we hope to design $f$ such that it is non-stationary with respect to \emph{all} finite-capacity agents $\pi$.

We note that for most environments studied in the literature, $f$ is non-stationary with respect to an \emph{unbounded} agent for which for all $t, U_t = H_t$.  Concretely, in these environments,
$$\limsup_{t\to\infty}\ \I(X_{t+1};X_{t+2:\infty}|H_t) > 0.$$
For instance, in an environment such as Permuted MNIST, there will continue to exist $t$ for which $X_{t+1}$ is data from a \emph{new} permutation.  As a result, this $X_{t+1}$ will contain information about the future sequence which is absent from $H_t$ and hence the environment is non-stationary even to an \emph{unbounded} agent.

However, we make the point that even if
\begin{equation}\label{eq:env_stationary}
    \limsup_{t\to\infty}\ \I(X_{t+1};X_{t+2:\infty}|H_t) = 0,  
\end{equation}
$f$ may be non-stationary with respect to $\pi$ due to capacity constraints.  Since the agent is only able to retain finite bits of information about the past, there will always be more to learn about in the future. We argue that environments which satisfy Equation \ref{eq:env_stationary} along with Definition~\ref{def:non-stationarity} better capture the essence of ``small agent, complex world.''  

Our notion of non-stationarity relates to the work of \cite{Abel2023definition}, which proposes a definition of continual reinforcement learning.  That definition offers a formal expression of what it means for an agent to never stop learning.  The characterization is subjective in that it defines non-convergence with respect to a {\it basis}, which is a fixed set of policies.  Definition \ref{def:non-stationarity} is similarly subjective, defining non-stationarity with respect to a particular agent with a particular agent state.

Given our definition of non-stationarity, we return to the implications on the behavior of an optimal finite capacity agent. An agent with finite capacity may need to ``forget'' some information in its agent state in order to incorporate new information, a concept which we elaborate on in Section~\ref{sec:forgetting_and_plasticity}. There may exist environments which are non-stationary but still exhibit a \emph{convergent} optimal policy.  This could be the case for problem settings in which the entire history is \emph{equally informative} about the future.  In such settings, ingesting the new information in $X_{t+1}$ and forgetting old information could be equivalent to ignoring $X_{t+1}$ entirely given capacity constraints.

However, such environments do not reflect reality nor do they align with our goals of developing effective continual learning algorithms.  To ensure that the environment adequately evaluates continual learning capabilities, it should be designed so that an optimal finite capacity agent \emph{never stops} ingesting (and forgetting) information.
Concretely, if we assume that the agent state $U_t$ is capacity constrained in that for all $t$, $\I(U_t;H_t) \leq c$, then the environment should require an optimal capacity constrained agent to never stop learning i.e.,
$$\limsup_{t\to\infty}\ \I(X_{t+1};U_{t+1}|U_t) > 0.$$
There will always be new information that the agent will need to incorporate into its agent state by forgetting old information.

\section{Forgetting and Implasticity}\label{sec:forgetting_and_plasticity}

In this section, we draw the connection between our notion of error $(\Lc_\pi)$ and phenomena described in the literature (notably ``forgetting'' and ``plasticity loss'' or ``implasticity'').  We begin by providing the following decomposition of error into forgetting and implasticity. This decomposition quantitatively improves upon the result of \cite{kumar2023continual} to provide definitions of forgetting and implasticity which better match our intuition. We defer the proof to Appendix~\ref{sec:appendix1}.
\begin{theorem}{\bf (forgetting and implasticity)}\label{th:forget_plast}
    For all agents $\pi:\mathcal{U}\times \Xc \mapsto \mathcal{U}$, if for all $t,\ U_{t+1} = \pi(U_t, X_t)$, then
    $$\Lc_{\pi}\ =\ \limsup_{T\to\infty}\ \underbrace{\frac{1}{T}\sum_{t=0}^{T-2}\I(X_{T:t+2};U_t|U_{t+1}, X_{t+1})}_{\rm forgetting} + \underbrace{\frac{1}{T}\sum_{t=0}^{T-1}\I\left(X_{T:t+1};X_t|U_t\right)}_{\rm implasticity}.$$
\end{theorem}
The forgetting that an agent experiences at time $t+1$ is the information about the future sequence $X_{T:t+2}$ which was contained in the previous agent state $U_t$, but not in the current agent state $U_{t+1}$ nor the most recent observation $X_{t+1}$.  Since agent state $U_{t+1} = \pi(U_t, X_{t+1})$, this mutual information exactly captures which \emph{relevant} information was decanted from the agent state incorporate information from $X_{t+1}$.  Note that forgetting is \emph{forward looking} as it measures mutual information between the previous agent state and the \emph{future} sequence as opposed to the \emph{past} sequence as done in the literature.  Forgetting about the past is not ``catastrophic'' insofar as what is forgotten is uninformative about the future.

Meanwhile, the implasticity that an agent experiences at time $t$ is the information about the future sequence $X_{T:t+1}$ which was contained in $X_t$, which we \emph{failed} to include in $U_{t}$.  This directly reflects what is referred to in the literature as ``loss of plasticity''.  An agent loses plasticity ,or incurs loss from implasticity, if it fails to digest information about $X_t$ which is relevant for making predictions about the future $X_{T:t+1}$.

Theorem \ref{th:forget_plast} suggests that if our environment is $k$-complex, then an effective continual learning agent will effectively reduce $\Lc_\pi$ by improving upon forgetting and/or implasticity with this additional compute.  Perhaps with this additional capacity, the agent will be able to hold on to more \emph{relevant} information throughout its experience and incur lower error due to not forgetting.  Or perhaps, the agent will elect to extract relevant information out of the data at each time step and suffer less error from implasticity. However, it's evident that a capacity-constrained agent in a $k$-complex environment will inevitably experience error due to forgetting and implasticity.

We now draw the connection to the criterion that a finite capacity agent should never stop learning.  Consider an agent $\pi$ which \emph{stops} learning i.e. for some $T$, $U_{t} = U_{T}$ for all $t > T$.  If the $f$ is non-stationary w.r.t $\pi$, then
\begin{align*}
    0 
    & \overset{(a)}{<} \I(X_{t+1};X_{t+2:\infty}|U_{t})\\
    & \overset{(b)}{=} \underbrace{\I(X_{t+1};X_{t+2:\infty}|U_{t+1})}_{\rm implasticity}
\end{align*}
where $(a)$ follows from the definition of non-stationarity and $(b)$ follows from the fact that $\pi$ stops learning.  Therefore, if $f$ appears non-stationary to all agents $\pi$, then an agent which \emph{stops learning} will \emph{necessarily} incur error due to implasticity.  To mitigate this, an effective capacity-constrained agent ought to never stop learning.

\section{An Illustrative Example: Turing-complete Prediction Environment}\label{sec:synthetic_env}

In this section, we propose a Turing-complete machine as an illustrative example of a big world simulator.
A Turing-complete machine can be considered the biggest possible world because it is capable of executing any computable program, world, or environment \citep{turing36_comput_number_applic_entsc}.
The specific machine that we use to simulate the big world is the cellular automaton Rule 110, which is the only elementary cellular automaton proven to be Turing-complete \citep{cook04_univer}. Figure~\ref{fig:rule110} in Appendix~\ref{appendix:more_ca} illustrates how Rule 110 updates each cell.

\begin{figure}[t]
    \centering
    \begin{minipage}[t]{0.48\linewidth}
        \centering
        \vspace{0.1mm}
        \includegraphics[width=0.49\linewidth]{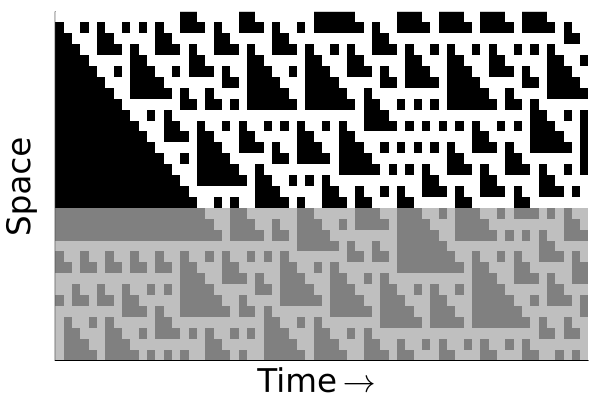}
        \includegraphics[width=0.49\linewidth]{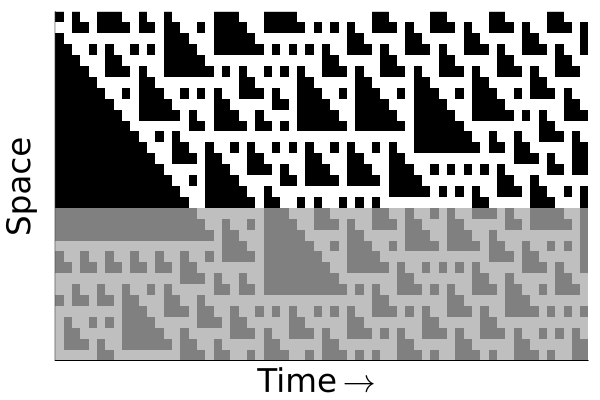}\\
        \includegraphics[width=0.49\linewidth]{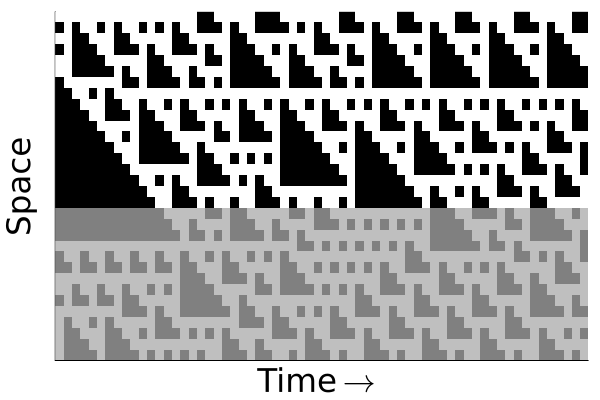}
        \includegraphics[width=0.49\linewidth]{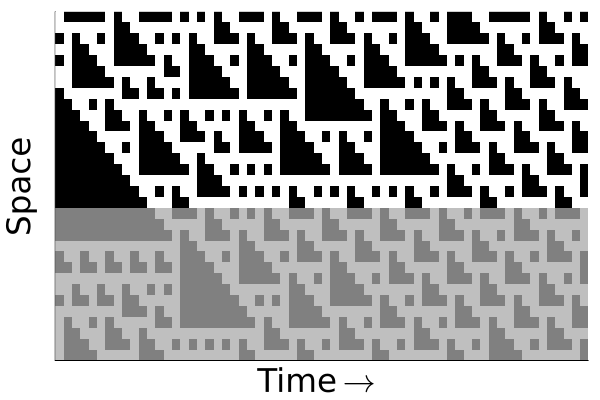}
        \caption{Four different simulations of Rule 110 with periodic boundary conditions, starting from the binary representation of integers 1 (top-left), 2 (top-right), 53 (bottom-left), and 107 (bottom-right). The gray shaded region is the unobserved region used to simulate the infinite state. At each time step, the agent observes a vertical slice of the unshaded region.}
        \label{fig:bigworld_example}
    \end{minipage}%
    \hfill
    \begin{minipage}[t]{0.48\linewidth}
        \centering
        \vspace{0.1mm}
        \includegraphics[width=0.99\linewidth]{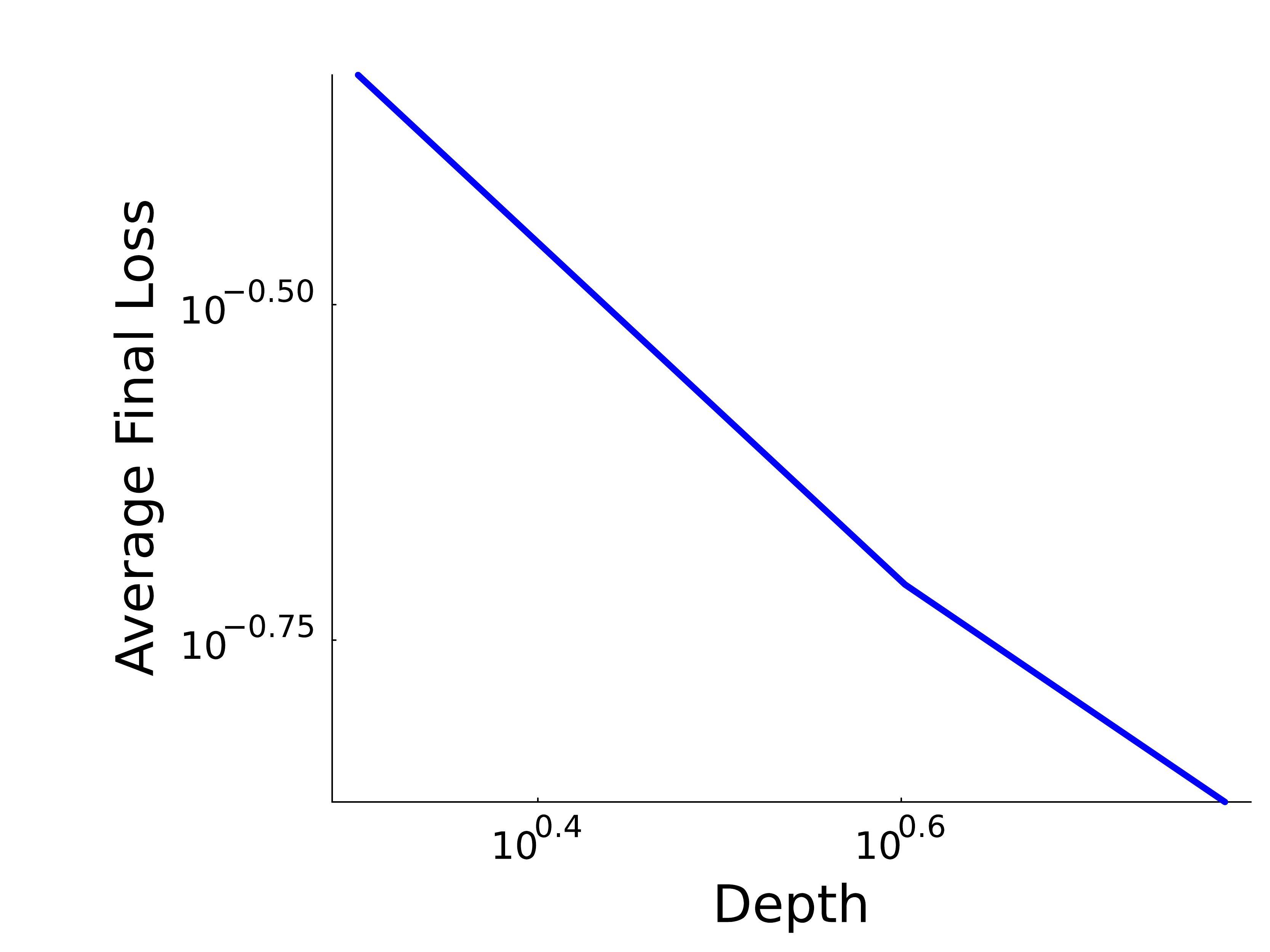}
        \caption{The error as a function of depth approximates $k$-complexity. Here, the depth of a feed-forward neural network is our measure of capacity.}
        \label{fig:scaling_result}
    \end{minipage}
\end{figure}

The two desiderata of a big world are satisfied in theory by the
the problem of predicting future observations from Rule 110. This is because Rule 110 defines a transition over an infinite state.
An agent with finite capacity tasked with predicting a future state can only observe a finite region of the state and can only predict up to some finite horizon.  Critically, the  region observed by the agent at each time step depends on both the observed and unobserved regions from previous time steps. Increasing the agent's capacity can improve its predictions, but the agent must never stop learning because of the infinite unobserved region.\footnote{To simulate the infinite state, we use a periodic boundary condition to connect the boundaries of the observed regions to the boundaries of the unobserved regions. While an agent with unbounded capacity could observe the infinite state, in our finite simulation, an agent with unbounded capacity could observe its entire history.} See Figure \ref{fig:bigworld_example} for different example simulations.

We now demonstrate that the Turing-complete prediction environment empirically satisfies the two desiderata of a big world simulator. We trained a neural network with SGD to predict the observation 16 steps into the future given the current observation.
In Figure \ref{fig:scaling_result}, we observe that increasing the agent's capacity (in the form of depth) leads to make increasingly more accurate predictions of the future observation. This relationship approximates $k$-complexity, and we anticipate a stronger relationship for the capacity of a history-encoding agent. Lastly, because of the unobserved region which simulates the infinite state, the agent must not stop learning. In Appendix \ref{appendix:more_ca}, we observed that some agent algorithms did stop learning due to loss of plasticity.

\section{Conclusion}

In this paper, we have highlighted the need for a big world simulator that accurately mirrors the challenges and complexities inherent in continual learning. We propose two properties which a big world simulator should exhibit: (1) there are no diminishing returns to increasing an agent's capacity, and (2) an optimal finite capacity agent should never stop learning.  Furthermore, we reiterate the concrete objective of minimizing average error over an infinite horizon subject to a capacity constraint and provide a decomposition of this objective into two terms which closely resemble the notions of ``forgetting'' and ``plasticity loss'' from the literature. Finally, we present a Turing-complete prediction environment as an illustration of a simulator with desired big world properties. We hope that this work stimulates discussion around next steps for designing a simulator that facilitates the discovery of effective continual learning agent design and evaluation metrics which better reflect the objective.

\bibliography{main}
\bibliographystyle{rlc}

\newpage

\appendix
\section{Additional Details on Existing Continual Learning Benchmarks}
\label{app:related}

In addition to Permuted MNIST, there are several other benchmarks that follow a similar recipe.
The Split MNIST benchmark
\citep{zenke17_contin,nguyen18_variat_contin_learn} divides the MNIST dataset
into tasks based on digit groupings, requiring the model to learn to distinguish
between different sets of numbers sequentially.
In the case of Incremental
CIFAR-100 \citep{lopez-paz17_gradien}, tasks are created by progressively
introducing new classes, testing the model's ability to incorporate new
knowledge without forgetting the previously acquired knowledge.
This approach to creating a continual
learning problem has also been applied to reinforcement learning, in which an
agent plays a sequence of games in the Arcade Learning Environment \citep{abbas23_loss_plast_contin_deep_reinf_learn}.

\subsection{Non-synthetic Continual Learning Benchmarks}
\label{app:nonsynth}
There are also continual learning problems in which the dataset was curated
specifically for the purposes of continual learning. In CoRE50, temporally
correlated images for object recognition are introduced corresponding either to
new classes or to new instances of an already encountered task
\citep{lomonaco17_core5}. Stream51 takes a similar approach, using temporally
sequenced frames from natural videos \citep{roady20_stream}. Lastly, there have
also been efforts to curate datasets with natural non-stationarities that may
better reflect the real-world challenges associated with continual learning,
such as WILDS \citep{koh21_wilds}, Wild-time \citep{yao22_wild_time} and
TemporalWiki \citep{jang2022temporalwiki}.

\newpage
\section{Proofs}
\label{sec:appendix1}

We provide the proof of Theorem 1 below.

\textbf{Theorem 1.}
For all agents $\pi:\mathcal{U}\times \Xc \mapsto \mathcal{U}$, if for all $t,\ U_{t+1} = \pi(U_t, X_t)$, then
$$\Lc_{\pi}\ =\ \liminf_{T\to\infty}\ \underbrace{\frac{1}{T}\sum_{t=0}^{T-2}\I(X_{T:t+2};U_t|U_{t+1}, X_{t+1})}_{\rm forgetting} + \underbrace{\frac{1}{T}\sum_{t=0}^{T-1}\I\left(X_{T:t+1};X_t|U_t\right)}_{\rm implasticity}.$$

\begin{proof}
    \begin{align*}
        \Lc_\pi
        & = \liminf_{T\to\infty}\ \E_{\pi}\left[\frac{1}{T}\sum_{t=0}^{T-1}\KL\left(\Pr(X_{t+1}\in\cdot|H_t)\ \|\ \Pr(X_{t+1}\in\cdot|U_t)\right)\right]\\
        & = \liminf_{T\to\infty}\ \frac{1}{T}\sum_{t=0}^{T-1}\I(X_{t+1};H_t|U_t)\\
        & = \liminf_{T\to\infty}\ \frac{1}{T}\sum_{t=0}^{T-1}\I(X_{t+1};U_{0:t}, H_t|U_t)\\
        & = \liminf_{T\to\infty}\ \frac{1}{T}\sum_{t=0}^{T-1}\left(\sum_{k=0}^{t-1}\I(X_{t+1};X_k,U_k|U_{t:k+1}, X_{t:k+1})\right) + \I(X_{t+1};X_t|U_t) \\
        & = \liminf_{T\to\infty}\ \frac{1}{T}\sum_{t=0}^{T-1}\left(\sum_{k=0}^{t-1}\I(X_{t+1;X_k}, U_k|U_{k+1}, X_{t:k+1})\right) + \I(X_{t+1};X_t|U_t) \\
        & \overset{(a)}{=} \liminf_{T\to\infty}\ \frac{1}{T}\sum_{t=0}^{T-1}\sum_{k=0}^{t-1}\ \I(X_{t+1};U_{k}|U_{k+1}, X_{t:k+1}) + \I(X_{t+1};X_k|U_{k}, U_{k+1}, X_{t:k+1})\\
        & \quad + \frac{1}{T}\sum_{t=0}^{T-1} \I(X_{t+1}, U_{t+1};X_t|U_t)\\
        & = \liminf_{T\to\infty}\ \frac{1}{T}\sum_{k=0}^{T-2}\I(X_{T:k+2};U_k|U_{k+1}, X_{k+1}) + \I(X_{T:k+2};X_{k}|U_{k}, U_{k+1}, X_{k+1})\\
        & \quad + \frac{1}{T}\sum_{t=0}^{T-1} \I(X_{t+1}, U_{t+1};X_t|U_t)\\
        & \overset{(b)}{=} \liminf_{T\to\infty}\ \frac{1}{T}\sum_{k=0}^{T-2}\I(X_{T:k+2};U_k|U_{k+1}, X_{k+1}) + \frac{1}{T}\sum_{k=0}^{T-1}\I\left(X_{T:k+1};X_k|U_k\right),
    \end{align*}
    where $(a)$ follows from the fact that $\I(U_{t+1};X_t|U_t, X_{t+1}) = 0$ and $(b)$ follows from the fact that $\I(U_{k+1};X_k|U_k, X_{T:k+1}) = 0$.
\end{proof}

\newpage

\section{Additional Experiments on Turing-compete Prediction Environment}
\label{appendix:more_ca}
\begin{figure}[h]
  \centering
\includegraphics[width=0.99\linewidth]{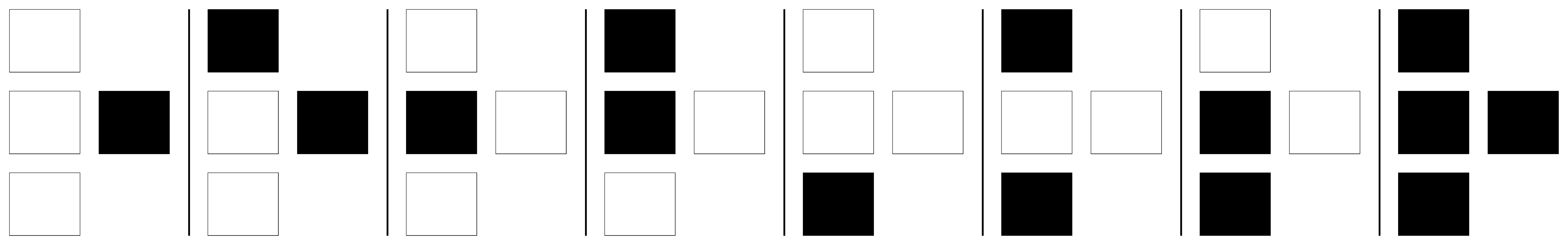}\\

\includegraphics[width=0.69\linewidth]{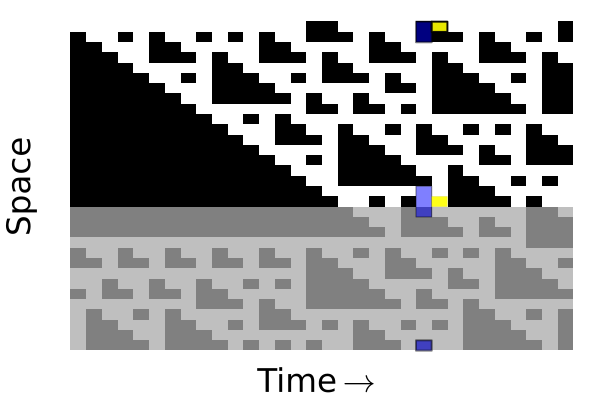}

    \caption{
    Top: Rule 110 updates each cell of the state using the value of the cell as well as its neighbours' values at the previous time step. (Left-Right): There are 8 possible configurations of the cell and its neighbours.
    Each configuration determines the cell value of the middle cell in the next state.
    Bottom: Zooming in on one simulation and the three cells in the blue box near the observable region, the fifth rule above is applied to output the middle cell's value at the next time step in the yellow box. This cell's value on the border of the observable region depends on the cell in the unobservable region. Similarly, the 2 blue cells at the top are connected to the single cell in the unobservable region in the bottom.
    }
  \label{fig:rule110}
\end{figure}

The Turing-complete Prediction Environment is defined by the 3-tuple, $\{S, O, K\}$. The state-size, $S$, is the size of the entire world. While the Turing-completeness of Rule 110 depends on an infinite initial state, we simulate this with a periodic boundary condition. The dynamics of the local state transition are shown in Figure \ref {fig:rule110}. The observation-size, $O$, controls the size of the observable world.
The unobserved region of the state space is depicted by the shaded region in Figure \ref{fig:bigworld_example}.
The prediction-horizon, $K$, controls the degree to which the non-stationarity induced by partial observability effects the prediction. This is because, for an elementary cellular automaton like Rule 110, each cell in the preceding state can only influence the cells to its left and right in the next state. Thus, the unobserved parts of the world can only influence the next state at the border of the observable region. The influence of the unobserved parts of the world on the prediction increases as the prediction-horizon increases.
In our experiments, we set the state space to be $32$ dimensional, the observation space to be the first $16$ dimensions and prediction horizons $K \in \{1, 2, 4, 8, 16\}$.

The environment is initialized to the state corresponding to the binary representation of $\tau = 1$, where $\tau$ is a starting state. 
Observations are provided sequentially for an episode length (or program length) of $T=100$.
After the $100$ steps of the program, the environment state is set to the binary representation of the next integer, $\tau \leftarrow \tau + 1$. We reset the state after a finite amount of steps to avoid the environment entering a loop of repeating states.
The programs for nearby integers is thus qualitatively similar, but varies for larger integers (see Figure \ref{fig:bigworld_example}).
Note that the starting state has a small initial state and an agent with unbounded capacity could encode the dynamics of the full state by compress the entire history

\begin{figure}[h]
  \centering

  \includegraphics[width=0.49\linewidth]{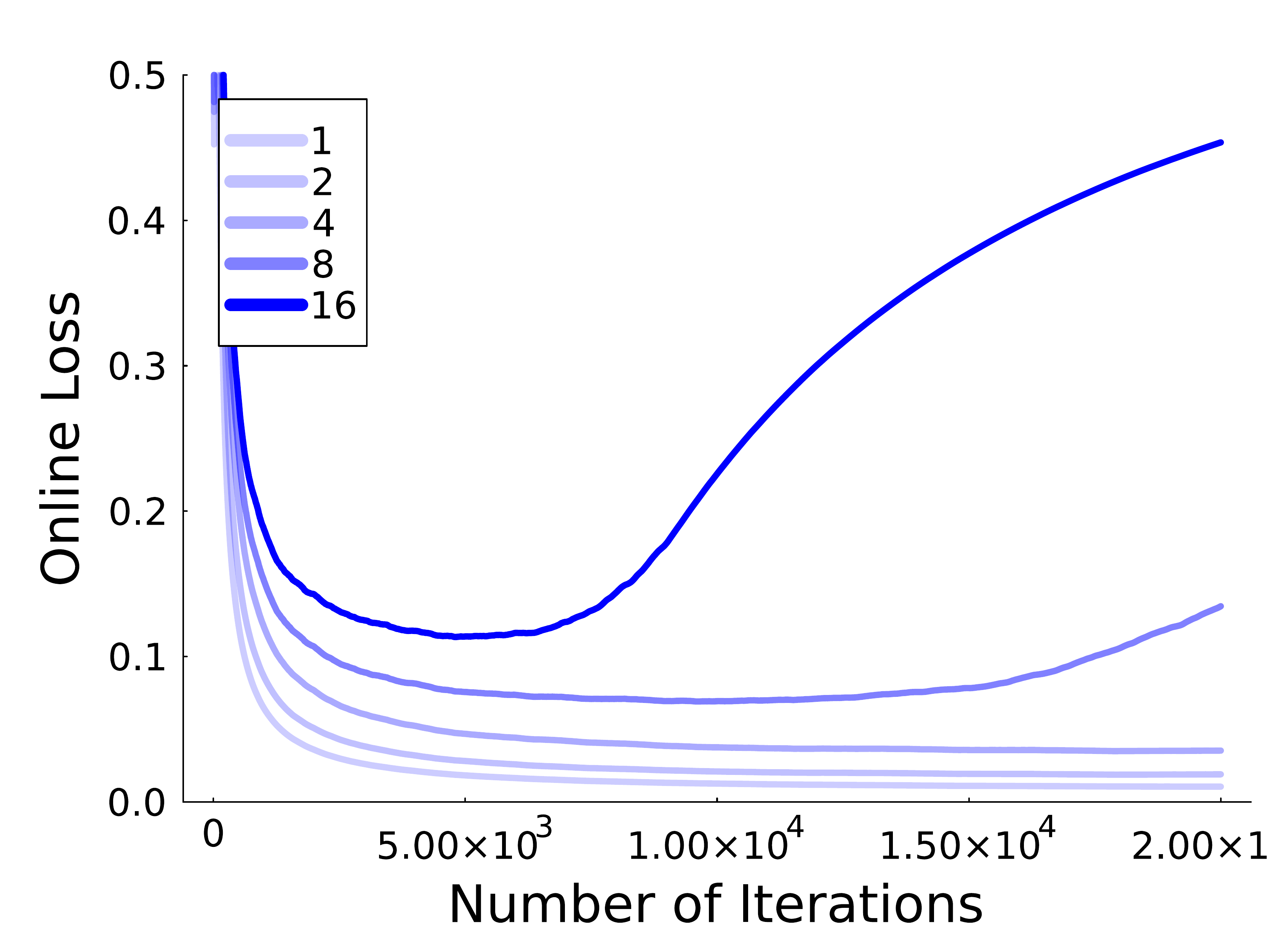}
  \includegraphics[width=0.49\linewidth]{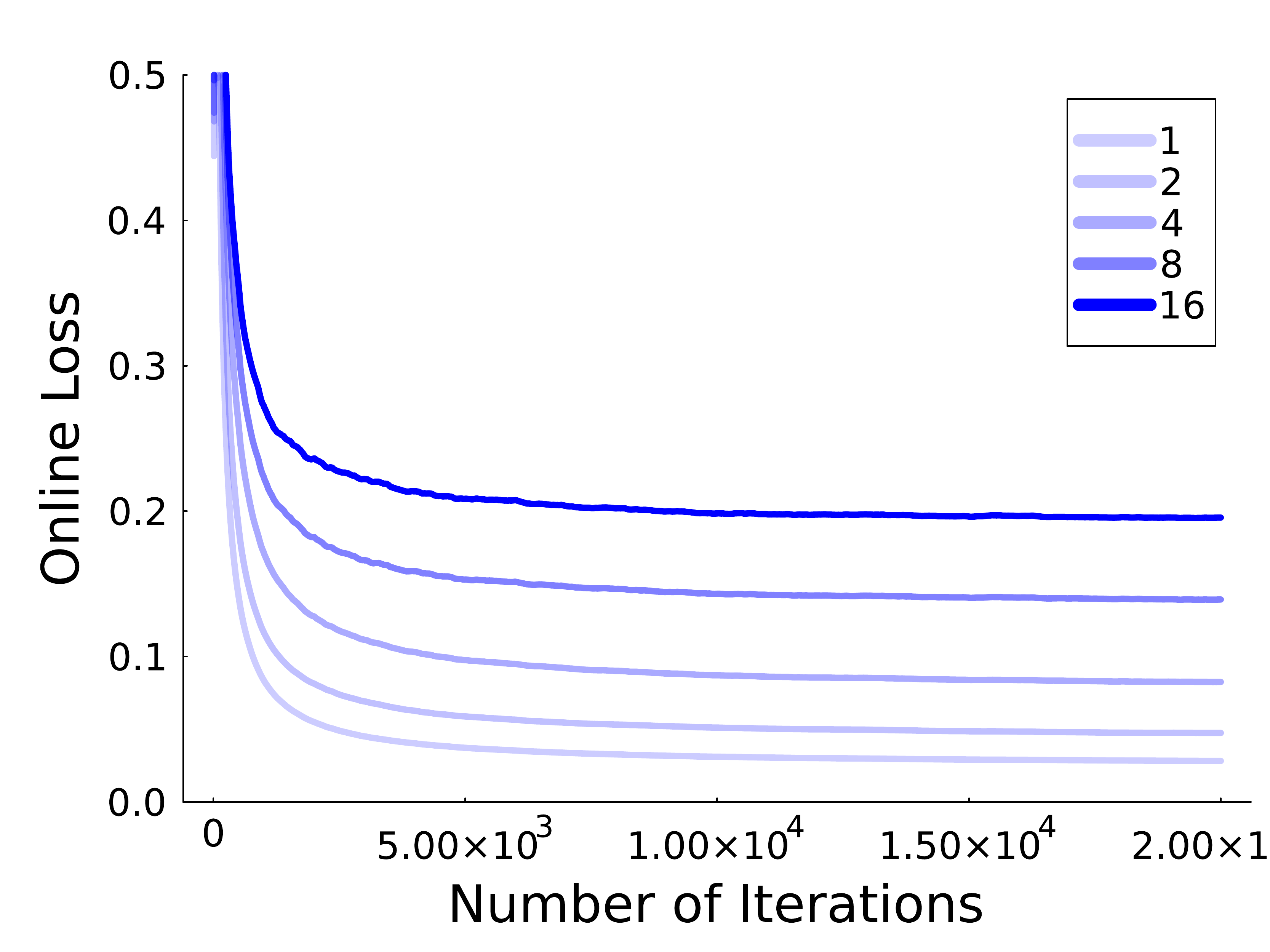}\\
  \includegraphics[width=0.49\linewidth]{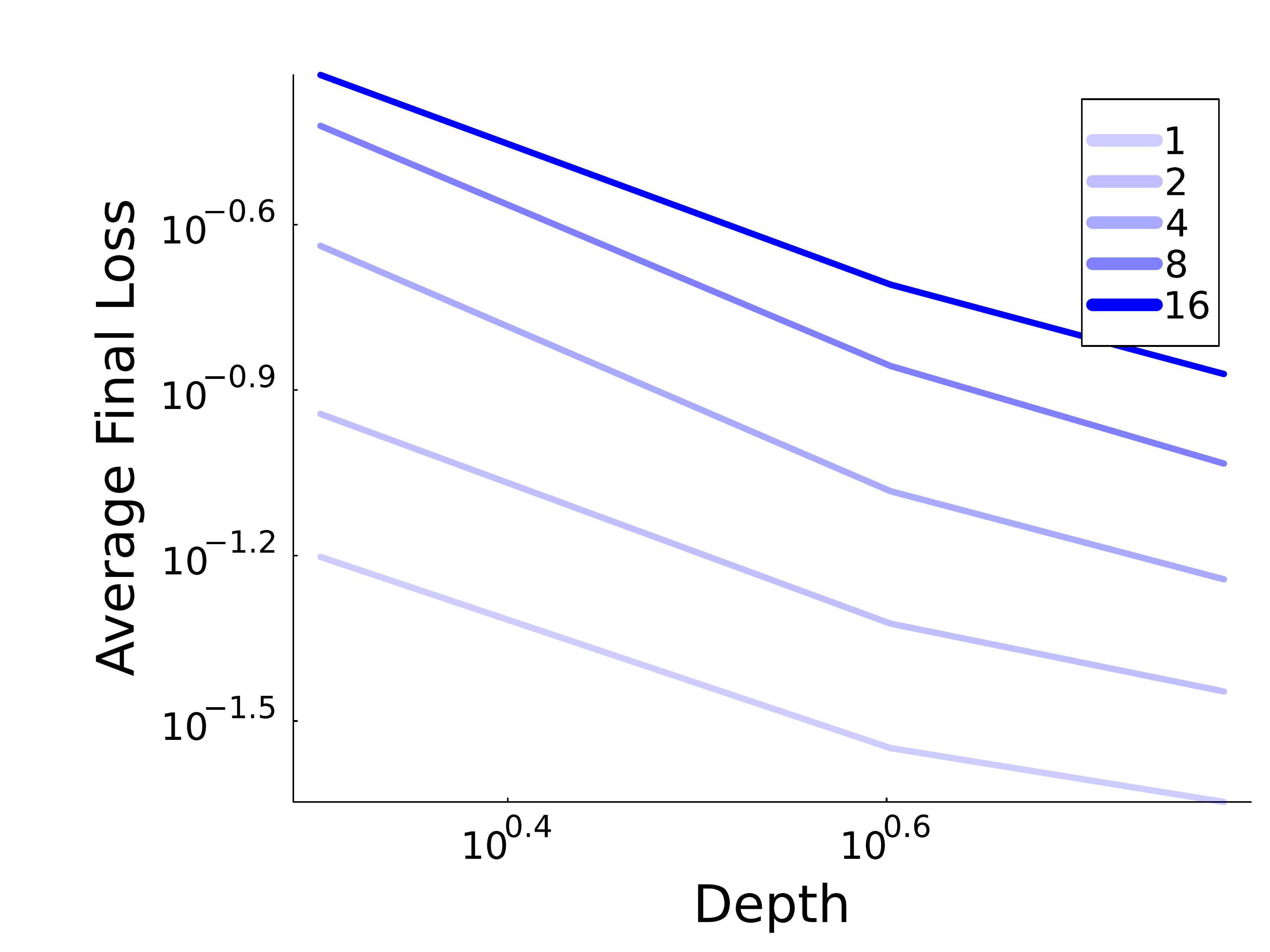}
  \includegraphics[width=0.49\linewidth]{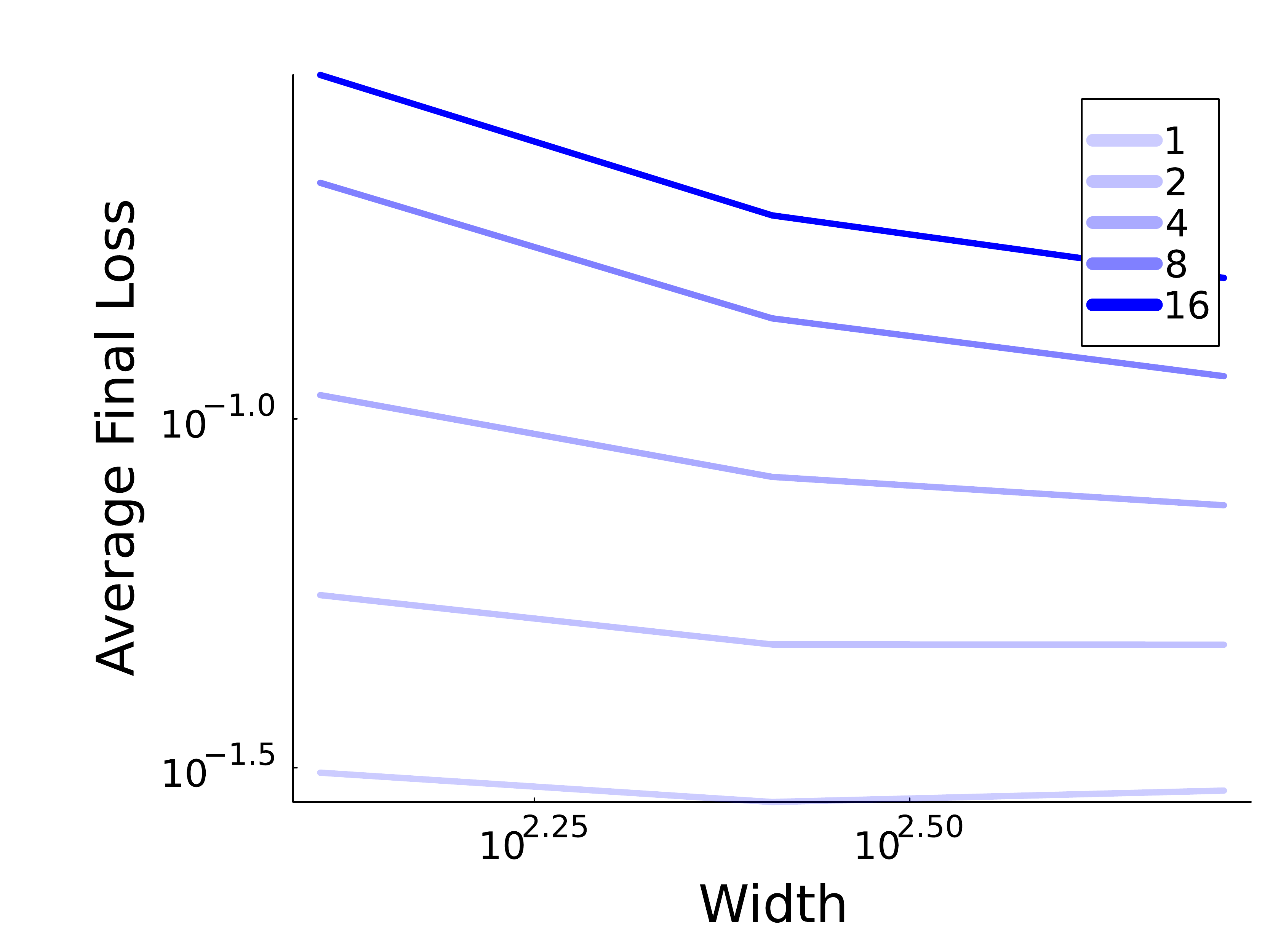}
  \caption{Top: Online accuracy for the medium sized neural network without regularization (left) and with regenerative regularization (right). Bottom: Larger prediction horizons are more difficult to make, but increasing capacity via depth (left) and width (right) leads to performance improvement. The opacity of each line indicates the prediction horizon.}
  \label{fig:main_results}
\end{figure}

The results are summarized in Figure \ref{fig:main_results}. For each prediction horizon given in the legend, we trained a neural network of different widths and depths to predict the state at the given horizon. On the top figure, we see that longer prediction horizons lead to more non-stationarity. The optimal agent must never stop learning but regularization towards the initialization is needed to sustain plasticity at longer prediction horizons~\citep{kumar23_maint_plast_regen_regul}. On the bottom figure, we see that doubling the capacity leads to approximately half the error at larger prediction horizons suggesting that this indeed simulates the big world properties that we have outlined.

\end{document}